\documentclass[12pt, a4paper]{article}

\usepackage[linesnumbered,ruled,vlined]{algorithm2e}
\usepackage{bm}
\usepackage{caption}
\usepackage{subfig}
\usepackage{amssymb}
\usepackage{authblk}
\usepackage[utf8]{inputenc}
\usepackage{hyperref}
\usepackage{subfig}
\usepackage{tikz}

\title{Estimating the concentration of gold nanoparticles incorporated 
on Natural Rubber membranes using Multi-Level Starlet Optimal 
Segmentation\footnote{\textit{Published in Journal of Nanoparticle
Research (December 2014).} The final publication is available at
\url{http://dx.doi.org/10.1007/s11051-014-2809-0}.}}

\author[a]{Alexandre Fioravante de Siqueira\footnote{Corresponding author. Phone: +55(19)3521--5362.\\
                                           \href{alexandredesiqueira@programandociencia.com}{alexandredesiqueira@programandociencia.com}}}
\author[b]{Flávio Camargo Cabrera\footnote{\href{flavioccabrera@yahoo.com.br}{flavioccabrera@yahoo.com.br}}}
\author[c]{Aylton Pagamisse\footnote{\href{aylton@fct.unesp.br}{aylton@fct.unesp.br}}}
\author[b]{Aldo Eloizo Job\footnote{\href{job@fct.unesp.br}{job@fct.unesp.br}}}

\affil[a]{DRCC -- Departamento de Raios Cósmicos e Cronologia,
          IFGW -- Instituto de Física ``Gleb Wataghin'',
          UNICAMP -- University of Campinas,
          Rua Sérgio Buarque de Holanda, 777, 13083-970,
          Campinas, São Paulo, Brazil}
\affil[b]{DFQB -- Departamento de Física, Química e Biologia,
          FCT -- Faculdade de Ciências e Tecnologia,
          UNESP -- Univ Estadual Paulista,
          Rua Roberto Simonsen, 305, 19060-900,
          Presidente Prudente, São Paulo, Brazil}
\affil[c]{DMC -- Departamento de Matemática e Computação,
          FCT -- Faculdade de Ciências e Tecnologia,
          UNESP -- Univ Estadual Paulista,
          Rua Roberto Simonsen, 305, 19060-900,
          Presidente Prudente, São Paulo, Brazil}

\begin{document}

\maketitle

\clearpage

\begin{abstract}
This study consolidates Multi-Level Starlet Segmentation (MLSS) and 
Multi-Level Starlet Optimal Segmentation (MLSOS), techniques for 
photomicrograph segmentation that use starlet wavelet detail levels to 
separate areas of interest in an input image. Several segmentation 
levels can be obtained using Multi-Level Starlet Segmentation; after 
that, Matthews correlation coefficient (MCC) is used to choose an 
optimal segmentation level, giving rise to Multi-Level Starlet Optimal 
Segmentation. In this paper, MLSOS is employed to estimate the 
concentration of gold nanoparticles with diameter around $47\,nm$, 
reducted on natural rubber membranes. These samples were used on the 
construction of SERS/SERRS substrates and in the study of natural rubber 
membranes with incorporated gold nanoparticles influence on 
\textit{Leishmania braziliensis} physiology. Precision, recall and 
accuracy are used to evaluate the segmentation performance, and MLSOS 
presents accuracy greater than 88\% for this application.

\vspace{0.2cm}
\end{abstract}

\section{Introduction} 

Increasingly researches are shifting their attention toward green 
chemistry, the use of eco-friendly substances for synthesizing 
nanostructures. Currently, green synthesis of gold nanoparticles 
is typically based on natural materials such as leaves 
\cite{ANNAMALAI2013,KUMAR2011}, fruits \cite{SUJITHA2013,GAJANAN2011}, 
seeds \cite{AROMAL2012} and flowers \cite{VIJAYAKUMAR2011,NORUZI2011}. 
Microbiological compounds from algae bacteria \cite{PARIAL2012} and 
fungi \cite{KANNAN2014} were also proposed. Gold nanoparticles 
synthesized by green chemistry have been used in different areas, as 
biosensors \cite{LIU2013}, antibacterial activity \cite{NAGAJYOTHI2014} 
and cancer treatments \cite{LIN2013}.

\cite{CABRERA2013A} has recently reported green synthesis of gold 
nanoparticles using natural rubber membranes from \textit{Hevea 
brasiliensis}, employed on the construction of flexible SERS and SERRS 
substrates \cite{CABRERA2012}, and also in the study of NR/Au 
influence on \textit{Leishmania braziliensis} physiology 
\cite{BARBOZAFILHO2012}. For a better understanding of the reduction 
kinetics of nanoparticles incorporated on solid substrates, image 
processing techniques can be used, such as object recognition and 
segmentation.

Gold nanoparticles in NR/Au samples were separated using starlet 
wavelet decomposition levels and choosing the optimal segmentation 
manually \cite{DESIQUEIRA2014A}, based in precision, recall and 
accuracy values. Enumerating the quantity of nanoparticles according 
to time reduction helps to understand the concentration of particles 
incorporated on solid substrates. However, this estimation is given 
manually, requiring operator intervention. This process could be 
facilitated by adopting an automatic method.

\subsection{Proposed approach}

In this paper we consolidate Multi-Level Starlet Segmentation (MLSS) 
and Multi-Level Starlet Optimal Segmentation (MLSOS) techniques, that 
receive assistance from Matthews correlation coefficient (MCC) in the 
detection of an optimal segmentation level \cite{DESIQUEIRA2014B}.
Using MLSS and MLSOS, we present here the concentration of gold 
nanoparticles incorporated on several natural rubber membranes through 
their photomicrographs, obtained from different magnifications. 

The proposed methodology consists of applying starlet wavelets in a 
NR/Au photomicrograph to acquire its detail decomposition levels. These 
levels are employed in MLSS, and the optimal segmentation level is 
obtained through MCC. After that, MLSOS can be applied in a set of 
photomicrographs, and the nanoparticle concentration is estimated. We 
use precision, recall and accuracy values in order to evaluate the 
performance of the segmentation, and MLSOS presents high accuracy in 
this application.

\section{Experimental} 

\subsection{Dataset of photomicrographs}

A dataset consisting of 30 images classified according to magnification, 
was employed to evaluate the proposed method. These images were obtained 
from natural rubber samples with gold nanoparticles using scanning 
electron microscopy (SEM). \cite{CABRERA2012,CABRERA2013A} presents the 
green synthesis of gold nanoparticles using natural rubber membranes, 
an organic component that acts as a reductant/stabilizer.

Gold nanoparticles were reduced in natural rubber at different time 
periods: 6, 9, 15, 30, 60 and 120 min. SEM images were obtained in 
magnifications of 25,000, 30,000, 100,000 and 200,000 times, using a 
FEI Quanta 200 FEG microscope with field emission gun (filament) 
equipped with a large field detector (Everhart-Thornley secondary 
electron detector), a solid state backscattering detector and pressure 
approx. of 1.00 Torr (low vacuum), as well as uncoated surface.
Photomicrographs in the dataset were obtained by secondary electron 
detectors.

\subsection{Starlet transform}

Starlet wavelet transform is an isotropic redundant wavelet based on 
the algorithm ``à trous'' (with holes) from \cite{HOLSCHNEIDER1990} and 
\cite{SHENSA1992}. These wavelets were successfully employed in analysis 
of astronomical \cite{STARCK2006,STARCK2010}, biological 
\cite{GENOVESIO2003} and microscopic 
\cite{DESIQUEIRA2014A,DESIQUEIRA2014B} images. Starlets are a good 
alternative in image processing and pattern recognition by their 
properties, such as isotropy, redundancy and translation-invariance.

Using the third order B-spline (B$_3$-spline) as its one-dimensional 
scale function, $\phi_{1D}$, the starlet transform is constructed as 
(Equations \ref{STARSCALE} and \ref{STARWAVE}, \cite{STARCK2011}):
\begin{eqnarray}
    \phi_{1D}(t) = \frac{1}{12}\left(|t-2|^3-4|t-1|^3+6|t|^3-4|t+1|^3+|t+2|^3\right) \label{STARSCALE} \\
    \frac{1}{2}\psi_{1D}\left(\frac{t}{2}\right) = \phi(t)-\frac{1}{2}\phi\left(\frac{t}{2}\right), \label{STARWAVE}
\end{eqnarray}
where $\psi_{1D}$ is the starlet one-dimensional analyzing wavelet.

A two-dimensional extension is achieved by a tensor product 
between the scale function and the analyzing wavelet (Equation 
\ref{STARLET2D}):
\begin{eqnarray}
    \phi(t_1,t_2) = \phi_{1D}(t_1)\phi_{1D}(t_2), \nonumber\\
    \frac{1}{4}\psi\left(\frac{t_1}{2},\frac{t_2}{2}\right) = \phi(t_1,t_2)-\frac{1}{4}\phi\left(\frac{t_1}{2},\frac{t_2}{2}\right). \label{STARLET2D}
\end{eqnarray}

Similarly to Equation \ref{STARLET2D}, the pair of finite impulse 
response (FIR) filters ($h, g = \delta - h$) related to this wavelet is 
(Equation \ref{FILTERPAIR}, \cite{STARCK2010}):
\begin{eqnarray}
    h_{1D}[k] = [\begin{array}{ccccc} 1 & 4 & 6 & 4 & 1\end{array}]/16, k = -2,...,2 \nonumber\label{H1D}\\ 
    h[k,l] = h_{1D}[k]h_{1D}[l] \nonumber\\
    g[k,l] = \delta[k,l]-h[k,l], \label{FILTERPAIR}
\end{eqnarray}
where $\delta$ is defined as $\delta[0, 0] = 1$, $\delta[k, l] = 0$ for 
$[k, l] \neq 0$. From Equations \ref{STARLET2D} and \ref{FILTERPAIR}, 
detail wavelet coefficients are obtained from the difference between the 
current and previous resolutions. The application of the starlet wavelet 
can be performed by a convolution between an input image $c_0$ and the 
FIR filter derived from $\psi$ (Equation \ref{H2D}, \cite{STARCK2010}),
\begin{center}
    \begin{equation}
        h = \frac{1}{16} \left[\begin{array}{c} 1 \\ 4 \\ 6 \\ 4 \\ 1 \end{array}
        \right] * \left[\begin{array}{ccccc} 1 & 4 & 6 & 4 & 1 \end{array}\right]\frac{1}{16}
        = \left[\begin{array}{ccccc} \frac{1}{256} & \frac{1}{64} & \frac{3}{128} & \frac{1}{64} & \frac{1}{256} \\
        \frac{1}{64} & \frac{1}{16} & \frac{3}{32} & \frac{1}{16} & \frac{1}{64} \\ 
        \frac{3}{128} & \frac{3}{32} & \frac{9}{64} & \frac{3}{32} & \frac{3}{128} \\
        \frac{1}{64} & \frac{1}{16} & \frac{3}{32} & \frac{1}{16} & \frac{1}{64} \\ 
        \frac{1}{256} & \frac{1}{64} & \frac{3}{128} & \frac{1}{64} & \frac{1}{256} \end{array}\right].
        \label{H2D}
    \end{equation}
\end{center}

The result of this convolution is a set of smooth coefficients 
corresponding to the first decomposition level, $c_1$. Detail wavelet 
coefficients of the first decomposition level are obtained from 
$w_1 = c_0 - c_1$. If $L$ is the last desired level, resolution levels 
can be calculated by:
\begin{eqnarray*}
    c_{j} = c_{j-1} * h, \\
    w_{j} = c_{j-1} - c_{j},
\end{eqnarray*}

\noindent where $j = 0,\ldots, L$ and the symbol $*$ denotes the 
convolution operation. The set $W = \{w_1, \ldots, w_L, c_L\}$ obtained 
by these operations is the starlet transform of the input image at level 
$L$.

\subsection{Multi-Level Starlet Segmentation (MLSS)}

\cite{DESIQUEIRA2014A} proposed a segmentation tool for photomicrographs 
based on starlets. This technique consists in:
\begin{itemize}
    \item applying the starlet transform in an input image $c_0$, 
    resulting in $L$ detail levels: $D_1, \ldots, D_L$, where $L$ is 
    the last desired resolution level;
    \item to ignore first and second detail levels ($D_1$, $D_2$), due 
    to the large amount of noise;
    \item third to $i$ detail levels are summed, and $c_0$ is subtracted 
    from this sum $(R_i = (D_3 + ... + D_i) - c_0)$, where $3 \leq i \leq L$.
\end{itemize}

$R_i$ is the starlet segmentation related to starlet level $i$. 
Then, the set $R_W = \{R_3, R_4, \ldots, R_L\}$ is the multi-starlet 
segmentation set related to $W$. This technique will be denominated as 
Multi-Level Starlet Segmentation (MLSS).

\subsection{Ground Truth (GT) and Matthews Correlation Coefficient 
(MCC): choosing the optimal decomposition level}
\label{SECGTMCC}

Regions of interest in an input image could be represented in a ground 
truth (GT). In this application, black areas represents the background, 
whereas white areas represents nanoparticles in the original image. The 
concepts of true positives (TP), true negatives (TN), false positives 
(FP) and false negatives (FN) could be established by comparing GT with 
a segmented image. According to a sample image, comparing its ground 
truth with a segmented image gives us TP, TN, FP and FN values as:
\begin{itemize}
    \item\textbf{TP:} pixels correctly labeled as gold nanoparticles.
    \item\textbf{FP:} pixels incorrectly labeled as gold nanoparticles.
    \item\textbf{FN:} pixels incorrectly labeled as background.
    \item\textbf{TN:} pixels correctly labeled as background.
\end{itemize}

Based on TP, FP, TN and FN, Matthews correlation coefficient (MCC, 
\cite{MATTHEWS1975}) was used to establish the optimal level for method 
application, also offering an evaluation of the segmentation correctness 
(Equation \ref{MCC}):
\begin{equation}
    MCC = \frac{TP*TN-FP*FN}{\sqrt{(TP+FN)(TP+FP)(TN+FP)(TN+FN)}} \label{MCC}
\end{equation}

MCC measures how variables tend to have the same sign and magnitude, 
where $1$, zero and $-1$ indicates perfect, random and imperfect 
predictions, respectively \cite{BALDI2000}.

\subsection{Multi-Level Starlet Optimal Segmentation (MLSOS)}

An extension to MLSS was proposed in \cite{DESIQUEIRA2014B}, which uses 
MCC to obtain automatic retrieval of the optimal starlet segmentation 
level for photomicrographs with similar features contained in a dataset. 
This extension will be denominated as Multi-Level Starlet Optimal 
Segmentation (MLSOS), and consists of:
\begin{itemize}
    \item applying MLSS in an input image $c_0$ for $L$ desired starlet 
    decomposition levels, obtaining $R_W = \{R_3, R_4 , \ldots, R_L\}$;
    \item comparing the elements of $R_W$ with the image GT and 
    obtaining TP, TN, FP and FN;
    \item calculating MCC (Equation \ref{MCC}) for TP, FP, TN and FN 
    obtained for each $R_i$ , with $3 \leq i \leq L$.
\end{itemize}

The optimal starlet segmentation level, $R_{i} opt$, is the one that 
returns the highest MCC value.

\subsection{Precision, recall and accuracy: evaluating the segmentation 
quality}

Besides MCC, that also gives a comparation between the input image and 
its GT, the following measures (also based in TP, TN, FP and FN) were 
used to evaluate the quality of the segmentation provided by MLSS and 
MLSOS (Equation \ref{PRECRECACC}, \cite{OLSON2008,WITTEN2011}):
\begin{itemize}
    \item\textbf{Precision.} it is the rate between the number of 
    retrieved pixels that are relevant (TP) and the total number of 
    pixels that are relevant (TP+FP).
    \item\textbf{Recall.} it is the rate between the number of retrieved 
    pixels that are relevant (TP) and the total number of pixels that 
    were retrieved (TP+FN).
    \item\textbf{Accuracy.} it is the overall success rate (percentage 
    of samples correctly classified).
\end{itemize}

Then, precision, recall and accuracy are defined as:
\begin{eqnarray}
    precision &=& \frac{TP}{TP+FP}\times 100\% \nonumber \\
    recall &=& \frac{TP}{TP+FN}\times 100\% \label{PRECRECACC} \\
    accuracy &=& \frac{TP+TN}{TP+TN+FP+FN}\times 100\% \nonumber
\end{eqnarray}

Precision represents retrieved pixels that are relevant; recall, on the 
other hand, means relevant pixels that were retrieved. Finally, accuracy 
returns the proportion of true retrieved results. 100\%, 0 and -100\% 
indicate perfect, random and imperfect predictions in each case, 
respectively.

\subsection{Estimation of gold nanoparticle amount within NR/Au samples}
\label{ESTIMNP}

\begin{figure}[hbt]
    \begin{center}
        \includegraphics{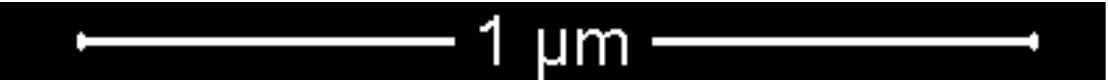}
    \end{center}
    \caption{Dimensional scale contained in SEM photomicrographs. This 
    bar represents $1 \mu m$, which in this example corresponds to 
    $400 px$.}
    \label{FIGDIMSCALE}
\end{figure}

After applying MLSOS in a photomicrograph of a NR/Au sample, the amount 
of gold nanoparticles on that sample could be estimated using the SEM 
information bar area with respect to sample size (Figure 
\ref{FIGDIMSCALE}). This bar has a dimensional scale measure that could 
be converted from length units to pixels.

Let $ratio_{lu}$ and $ratio_{px}$ be the value of length units in the 
photomicrograph dimensional scale and its equivalent in pixels, 
respectively. Assuming that the total amount of nanoparticles is 
represented by the total of white pixels in the input image, 
nanoparticles from the segmented image could be estimated from Equation 
\ref{EQQTTY}:
\begin{equation}
    total_{au} = \left(\frac{ratio_{lu}}{ratio_{px}}*total_{NP}\right)^2, \label{EQQTTY}
\end{equation}
where $total_{au}$ is the nanoparticle concentration in an area unit.
Algorithms for MLSS aimed to these photomicrographs are presented in
\cite{DESIQUEIRA2014A}. Also, the source code in Octave programming 
language\footnote{GNU Octave is an open source high-level interpreted 
language intended primarily for numerical computation. Download 
available freely at http://www.gnu.org/software/octave/download.html.} 
for starlet transform application and MLSOS computation is available 
in \cite{DESIQUEIRA2014B}.

\section{Results and discussion} 

\begin{figure}[hbt]
    \begin{center}
        \includegraphics[width=0.6\textwidth]{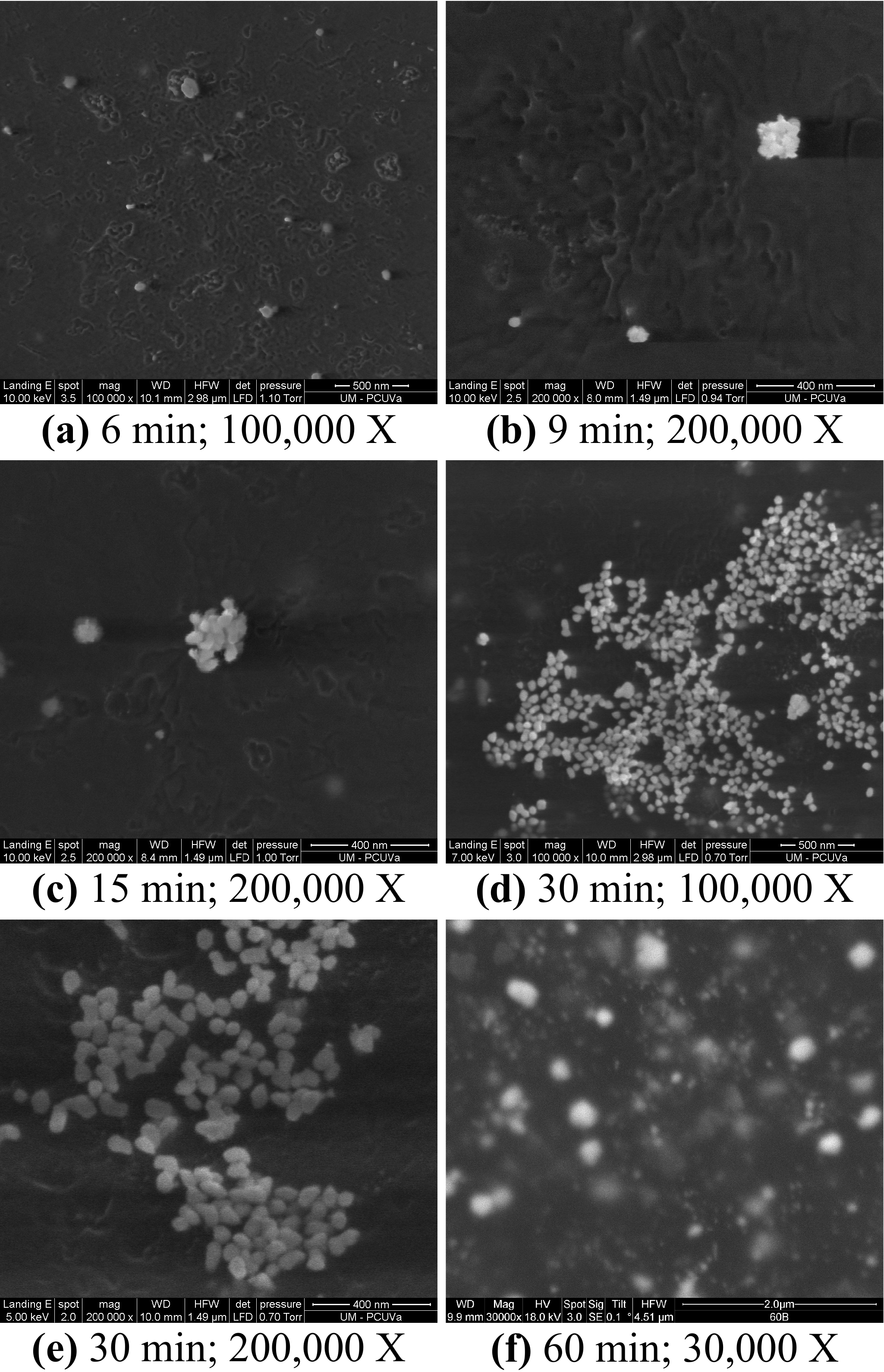}
    \end{center}
    \caption{Photomicrographs of natural rubber samples with 
    incorporated gold nanoparticles obtained by scanning electron 
    microscopy. Reduction time of the nanoparticles in the sample is 
    given in each Figure, as well as the magnification factor.}
    \label{FIGPHOTO}
\end{figure}

In order to obtain nanoparticle amount estimation, photomicrographs 
belonging to the dataset were classified in five sets, according to 
their magnification: 
$13,000\times$; $25,000\times$; $30,000\times$; $100,000\times$; 
$200,000\times$. Each one of these sets contains six photomicrographs, 
related to the reduction time of nanoparticles (6, 9, 15, 30, 60 and 
120 minutes).
The optimal segmentation level must be defined for MLSOS application. 
For this purpose, six test images from the dataset (Figure 
\ref{FIGPHOTO}), with their respective ground truths, were processed
using MLSS.

Figure \ref{FIGPHOTO}(a) exemplifies MLSS application using the last 
starlet resolution level $L = 10$. Thus, MLSS generates seven 
segmentation levels ($R_w = \{R_3,\ldots, R_{10}\}$) (Figure 3). Gold 
nanoparticles appear in white, whereas background is shown in black. 
As $R$ increases, areas attributed to gold nanoparticles by MLSS tend 
to grow and merge, giving rise to larger regions. Segmentation levels 
$R_6$, $R_7$ and $R_8$ have a better visual representation of 
nanoparticles presented in this photomicrograph; lower levels contains 
fewer nanoparticle information, whereas in higher levels regions 
assigned to nanoparticles are fused, leading to misinterpretation of 
material surface.
\begin{figure}[hbt]
    \begin{center}
        \includegraphics[width=0.6\textwidth]{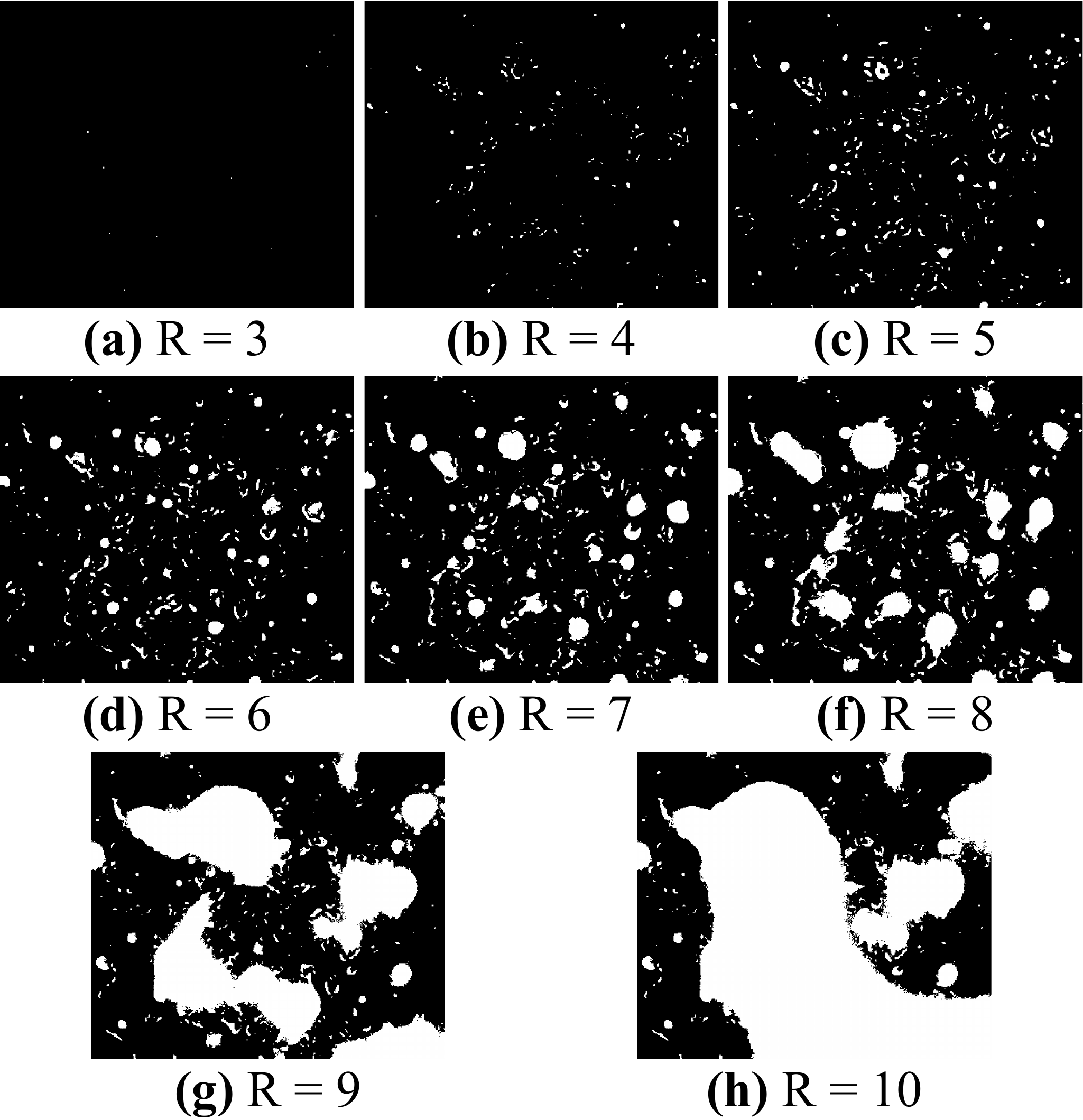}
    \end{center}
    \caption{Results of Multi-Level Starlet Segmentation (MLSS) for 
    Figure \ref{FIGPHOTO}(a). Starlet lower levels represent smaller 
    regions precisely. As the application level $R$ increases areas 
    detected by MLSS tend to merge, creating larger regions.}
    \label{FIGSTARAPPL}
\end{figure}

\begin{figure}[hbt]
    \begin{center}
        \includegraphics[width=0.6\textwidth]{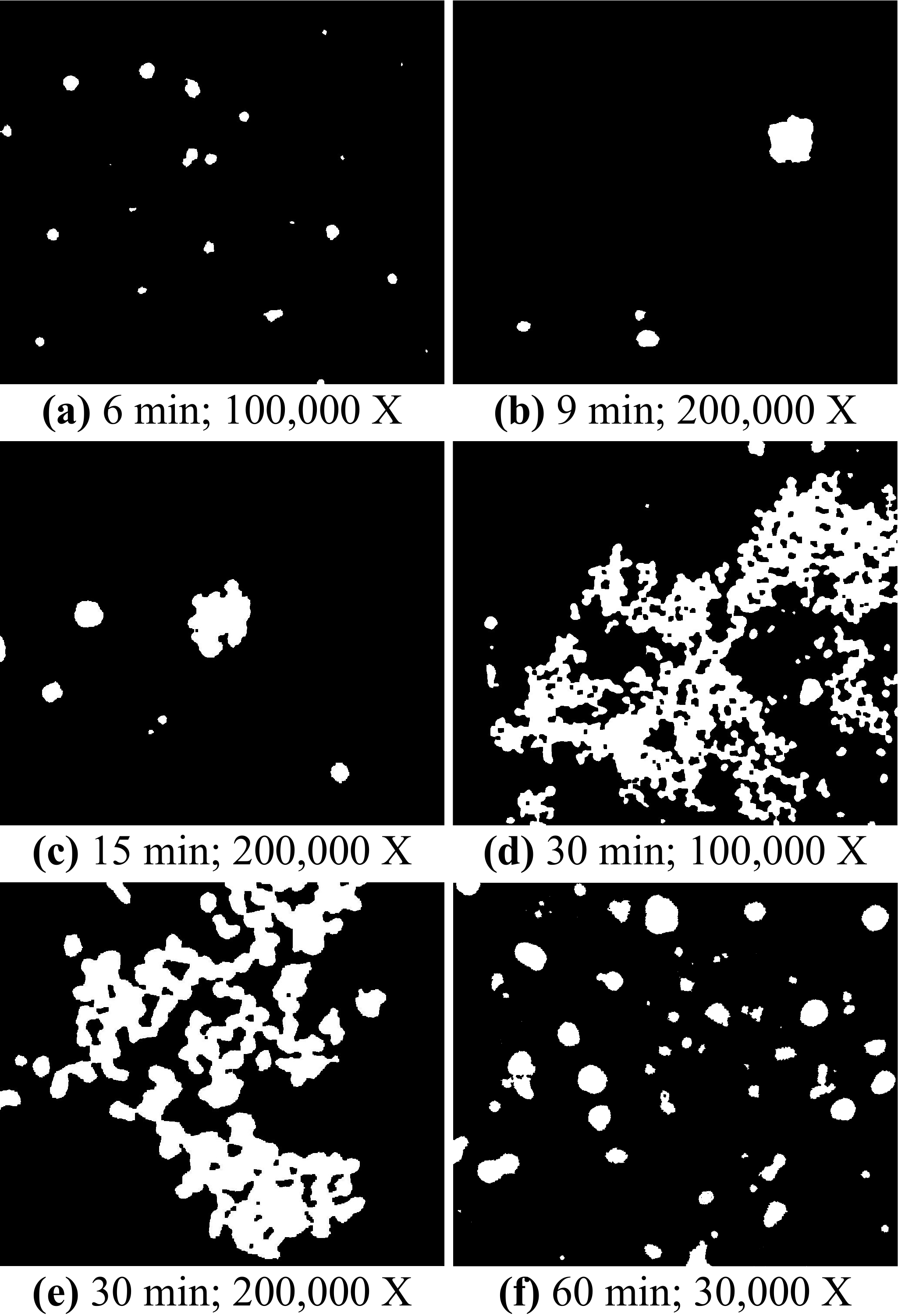}
    \end{center}
    \caption{Ground Truth (GT) of Figure \ref{FIGPHOTO} 
    photomicrographs. These images were obtained by a specialist, and 
    represent gold nanoparticles in a sample surface.}
    \label{FIGGT}
\end{figure}

After MLSS application, each $R_W$ element is compared to its 
corresponding GT\footnote{Ground truths were obtained by a specialist 
using GIMP (GNU Image Manipulation Program), a powerful open source 
graphics software. Its download is available for several operational 
systems at http://www.gimp.org/downloads.} (Figure \ref{FIGGT}) thus 
acquiring TP, FP, FN and TN (Section \ref{SECGTMCC}). MCC (Equation 
\ref{MCC}) is calculated for all $R_W$ elements (Table \ref{TABMCC}) 
from these parameters, and their values were compared. This comparison 
will determine the optimal segmentation level.

\begin{table*}[hbt]
    \caption{MCC obtained from Equation \ref{MCC} application in NR/Au 
    samples (Figure \ref{FIGPHOTO}) for levels $R_3$ to $R_{10}$. Higher 
    MCC values for each photomicrography are shown in red. $R_7$ was 
    chosen as the optimal segmentation level for Figure \ref{FIGPHOTO} 
    images, according to MCC values.}
    {\tiny
    \begin{center}
        \begin{tabular}{l*{8}{c}}
            \hline
            & \multicolumn{8}{c}{\textbf{MCC (\%)}} \\
            \hline
            \textbf{Reduction time (min); Magnification ($\bm{\times}$)} & \bm{$R_3$} & \bm{$R_4$} & \bm{$R_5$} & \bm{$R_6$} & \bm{$R_7$} & \bm{$R_8$} & \bm{$R_9$} & \bm{$R_{10}$}\\
            \hline
            \textbf{6; $\bm{100,000}$} & 5.342 & 22.382 & 32.335 & \color{red}{36.172} & 34.039 & 27.096 & 17.031 & 10.646 \\
            \textbf{9; $\bm{200,000}$} & 2.302 & 25.040 & 30.712 & 39.613 & \color{red}{49.271} & 47.863 & 34.236 & 18.007 \\
            \textbf{15; $\bm{200,000}$} & 7.642 & 31.534 & 49.005 & 65.083 & \color{red}{72.195} & 63.440 & 48.205 & 28.944 \\
            \textbf{30; $\bm{100,000}$} & 15.662 & 39.251 & 46.976 & 55.753 & 63.359 & \color{red}{67.580} & 65.125 & 59.147 \\
            \textbf{30; $\bm{200,000}$} & 4.314 & 32.183 & 53.876 & 65.137 & \color{red}{72.593} & 70.222 & 65.261 & 52.300 \\
            \textbf{60; $\bm{30,000}$} & 0 & 8.841 & 26.021 & 42.144 & 51.235 & \color{red}{56.591} & 45.687 & 33.489 \\
            \hline
        \end{tabular}
    \end{center}
    }
    \label{TABMCC}
\end{table*}

Best MCC values for Figure \ref{FIGPHOTO} images are given for 
segmentation levels $R_6$ (one time), $R_7$ (three times), and $R_8$ 
(two times); furthermore, the difference of sixth/eighth segmentation 
levels with higher MCC, when compared to $R_7$, lies between 
$2.1 \sim 5.4\%$. Based on these facts, $R_7$ is elected the optimal 
level for MLSOS application in dataset photomicrographs. 
In order to confirm $R_7$ as the optimal segmentation level, precision, 
recall and accuracy values for $R_6$, $R_7$ and $R_8$ were obtained and 
compared (Table \ref{TABPRA}). It can be seen that precision and recall 
values tend to decrease and increase, respectively, as segmentation 
levels become higher.

\begin{table*}[hbt]
    \caption{Precision, recall and accuracy values (Equation 
    \ref{PRECRECACC}) obtained by applying the proposed methodology in 
    Figure \ref{FIGPHOTO} fotomicrographs (segmentation levels $R_6$, 
    $R_7$, $R_8$). Higher values for accuracy, for each image, are 
    given in red.}
    {\tiny
    \begin{center}
        \begin{tabular}{l*{9}{c}}
            \hline
            & \multicolumn{3}{c}{\textbf{Precision (\%)}} & \multicolumn{3}{c}{\textbf{Recall (\%)}} & \multicolumn{3}{c}{\textbf{Accuracy (\%)}}\\
            \hline
            \textbf{Reduction time (min); Magnification ($\bm{\times}$)} & \bm{$R_6$} & \bm{$R_7$} & \bm{$R_8$} & \bm{$R_6$} & \bm{$R_7$} & \bm{$R_8$} & \bm{$R_6$} & \bm{$R_7$} & \bm{$R_8$}\\
            \hline
            \textbf{6; $\bm{100,000}$} & 86.639 & 49.675 & 16.970 & 52.425 & 88.695 & 97.446 & \color{red}{99.188} & 98.525 & 93.011 \\
            \textbf{9; $\bm{200,000}$} & 26.887 & 28.491 & 24.093 & 62.133 & 89.018 & 99.750 & \color{red}{96.983} & 96.581 & 95.412 \\
            \textbf{15; $\bm{200,000}$} & 65.648 & 54.820 & 41.941 & 66.385 & 97.393 & 99.746 & \color{red}{98.184} & 97.798 & 96.325 \\
            \textbf{30; $\bm{100,000}$} & 94.318 & 86.833 & 77.332 & 55.154 & 74.186 & 90.388 & 84.813 & 88.314 & \color{red}{88.616} \\
            \textbf{30; $\bm{200,000}$} & 85.822 & 81.991 & 67.341 & 62.612 & 78.389 & 94.300 & 86.493 & \color{red}{89.012} & 85.445 \\
            \textbf{60; $\bm{30,000}$} & 36.733 & 39.366 & 39.066 & 62.183 & 80.243 & 95.363 & 88.808 & \color{red}{88.929} & 88.155 \\
            \hline
        \end{tabular}
    \end{center}
    }
    \label{TABPRA}
\end{table*}

$R_6$ and $R_7$ are the levels with higher accuracy values. In cases 
where accuracy is higher for $R_6$, the difference compared with their 
respective $R_7$ values is lower than $0.7\%$. Therefore $R_7$ is a 
suitable choice of the optimal level.

\begin{figure}[hbt]
    \begin{center}
        \includegraphics[width=0.6\textwidth]{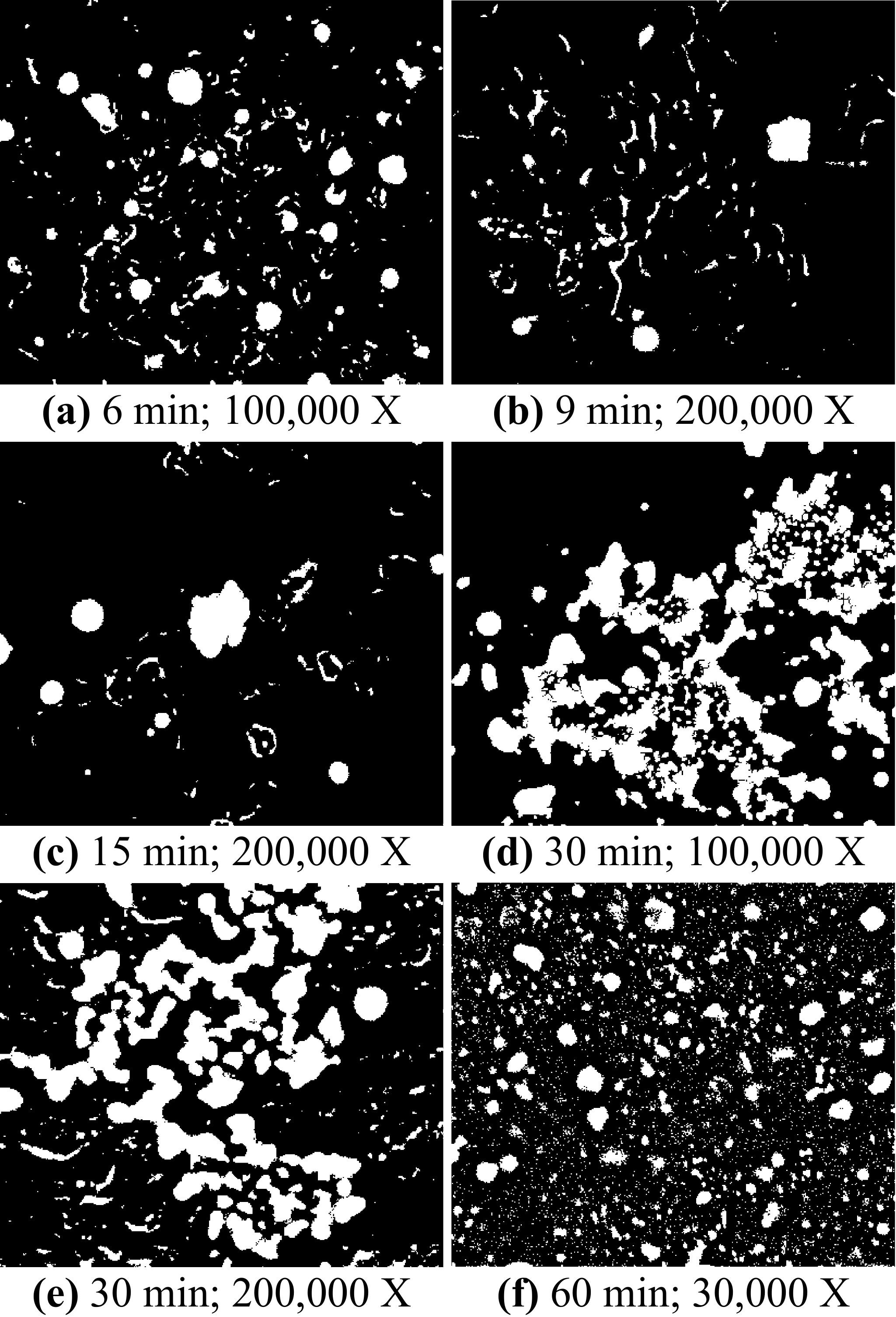}
    \end{center}
    \caption{Result of Multi-Level Starlet Optimal Segmentation (MLSOS) 
    application in Figure \ref{FIGPHOTO} photomicrographs. Optimal 
    level: $R_7$.}
    \label{FIGMLSOS}
\end{figure}

After choosing the optimal segmentation level, dataset images were 
processed using MLSOS. An example of the obtained segmentation using 
$R_7$ is given for Figure \ref{FIGPHOTO} photomicrographs (Figure 
\ref{FIGMLSOS}).

\begin{figure}[hbt]
    \begin{center}
        \includegraphics[width=0.6\textwidth]{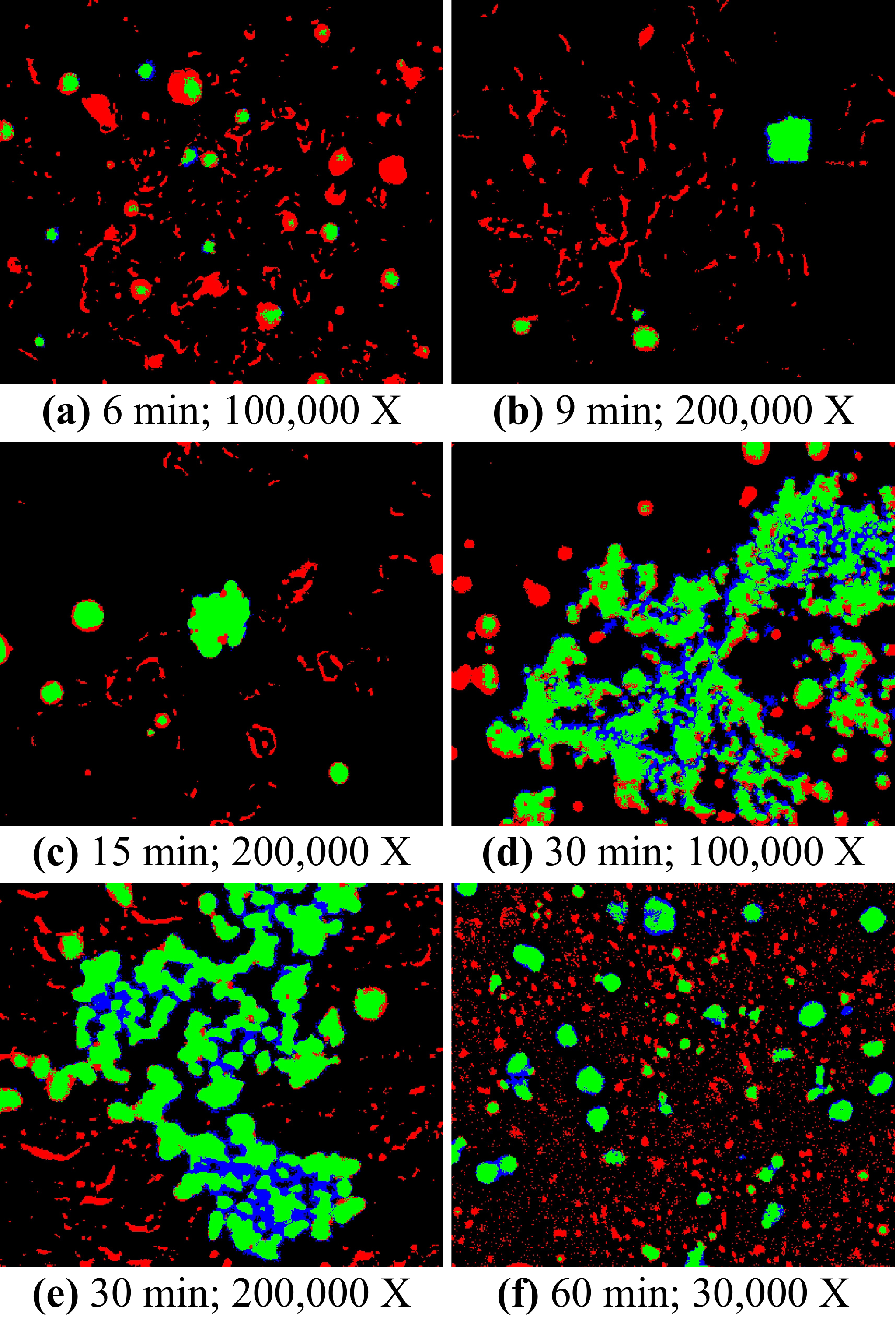}
    \end{center}
    \caption{Comparison between ground truths (Figure \ref{FIGGT}) and 
    Multi-Level Starlet Optimal Segmentation (MLSOS, Figure 
    \ref{FIGMLSOS}) for photomicrographs of Figure \ref{FIGPHOTO}. 
    Green: TP pixels; blue: FN pixels; red: FP pixels. Shapeless red 
    areas are attributed to defects in the material surface.}
    \label{FIGCOMP}
\end{figure}

A visual comparison between Figure \ref{FIGPHOTO} ground truths (Figure 
\ref{FIGGT}) and the result of MLSOS application (Figure \ref{FIGMLSOS}) 
is based on TP, FP, FN and TN (Figure \ref{FIGCOMP}), where green, blue 
and red represent TP, FN and FP pixels, respectively. MLSOS defines some 
surface defects of the material as areas with nanoparticles; such 
behavior is represented by shapeless red areas in this comparison.
Apparently, $R_7$ is not sufficient for segmenting nanoparticles 
contained in larger sets. This phenomenon is shown by the concentration 
of FN pixels (blue regions) appearing mostly contained in sets of TP 
pixels (green areas).

Since MLSOS results were acquired, Equation \ref{EQQTTY} can be applied 
for the estimation of nanoparticles in photomicrographs (Section 
\ref{ESTIMNP}, Figure 7). Each set was separated according to its 
magnification, containing photomicrographs related to the reduction 
time. Thereby the behavior of incorporated gold nanoparticles in a 
rubber sample can be evaluated according to reduction time.
\begin{figure}[hbt]
    \begin{center}
        \includegraphics[width=1\textwidth]{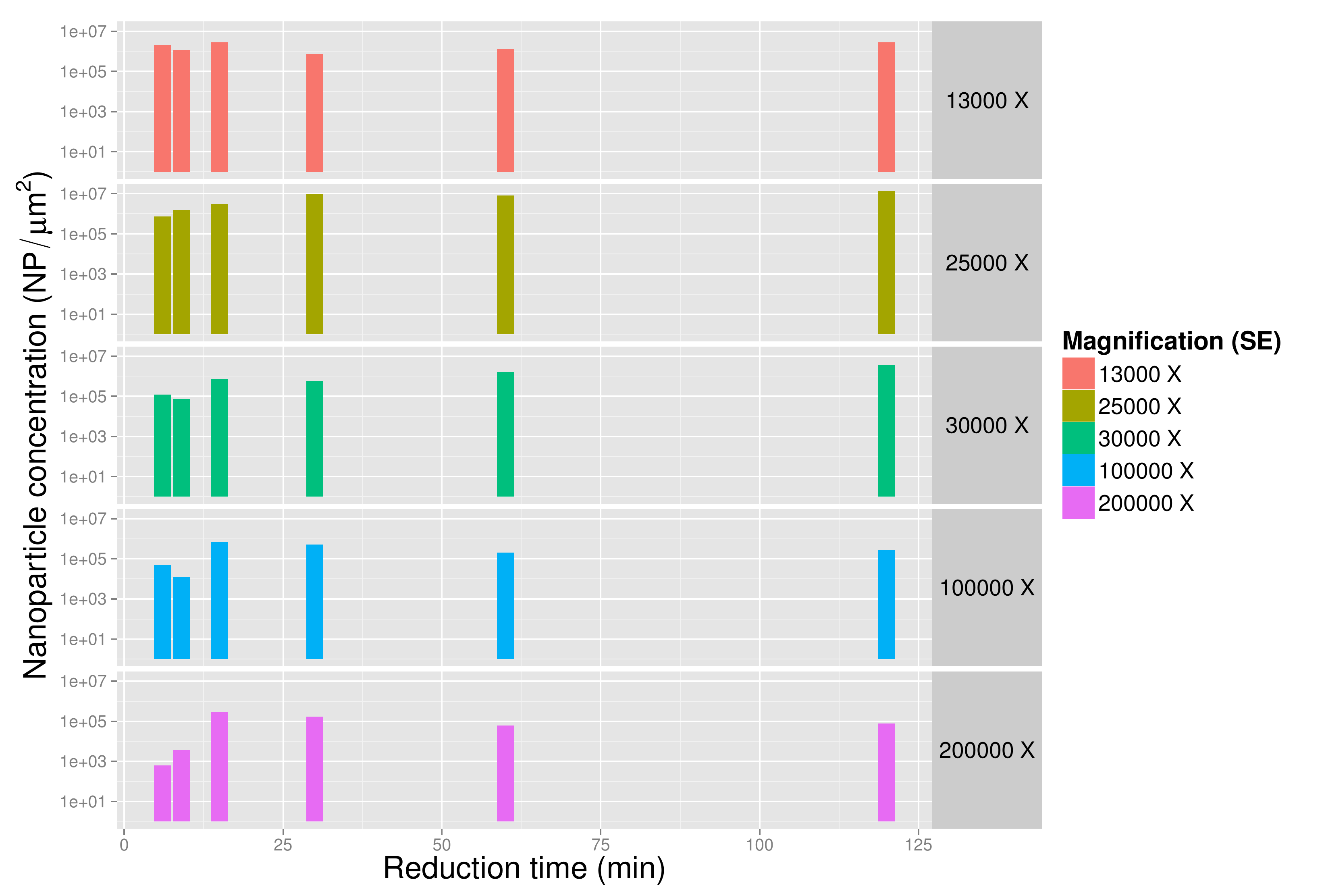}
    \end{center}
    \caption{Area attributed to gold nanoparticles using Multi-Level 
    Starlet Optimal Segmentation (MLSOS) in $13,000\times$ to 
    $200,000\times$ image sets ($R_7$, secondary detector). The bars 
    represent the nanoparticle amount by $\mu m^2$ for different gold 
    reduction times, namely: 6, 9, 15, 30, 60 and 120 minutes, for each 
    magnification.}
    \label{FIGQTTY}
\end{figure}

\begin{figure}[hbt]
    \begin{center}
        \includegraphics[width=1\textwidth]{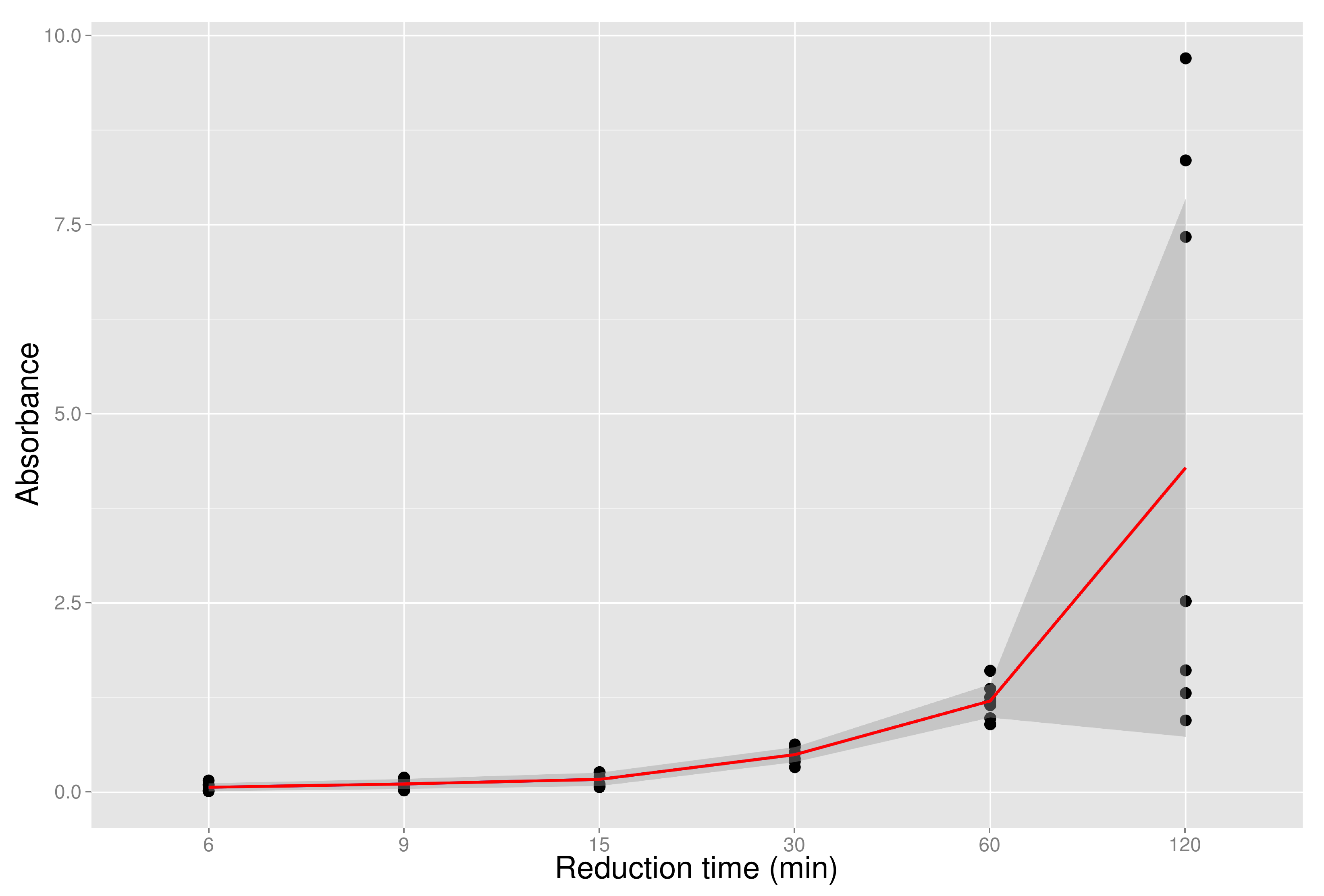}
    \end{center}
    \caption{Temporal evolution from incorporation of gold nanoparticles 
    on the surface of natural rubber membranes, measured in terms of 
    increase in the maximum of the absorption band due to reduction 
    time. Black dots: NR/Au samples absorbance. Red line: mean 
    evolution. Shaded area: standard deviation evolution.}
    \label{FIGUVVIS}
\end{figure}

MLSOS results are compared to temporal evolution of nanoparticles 
incorporation on the surface of natural rubber samples, measured in 
terms of increase in the maximum of the plasmon absorption band due to 
reduction time. These measures were performed as a comparative basis 
for membranes prepared from latex obtained from different collections, 
conducted at different times of the year, evaluated by ultraviolet-visible 
spectroscopy (UV-Vis, Figure \ref{FIGUVVIS}), where black dots 
represent NR/Au samples absorbance. The red line and the shaded area 
presents, respectively, mean and standard deviation evolution between 
samples with same reduction time.

As shown in UV-Vis measures, temporal evolution presents an increase 
depending on the reduction time. This increase in intensity of 
absorption bands is given by the rise in the amount of gold 
nanoparticles deposited on natural rubber surface, that could generate 
nanoparticle aggregates mainly observed after the reduction time of 
30 minutes \cite{CABRERA2013B}. Reliable reproducibility of results 
is given up to the reduction time of 60 minutes, given the small 
standard deviation obtained from comparison of different samples. These 
results indicate also a reduction timeout of 120 minutes, whereas some 
samples exceeds absorbance measuring limit of the equipment due to the 
formation of a thick layer of nanoparticles and aggregated bulk, 
decreasing the reproducibility of results.

Nevertheless, it is necessary to remark that electron microscopy only 
evaluates nanoparticles in the surface of natural rubber membranes, 
whereas UV-Vis spectroscopy evaluates all deposited nanoparticles 
between the optical analysis path, since it was previously demonstrated 
that particles could be incorporated within 3 micrometers in the volume 
of natural rubber membranes \cite{CABRERA2013A}.

According to nanoparticle evaluation through MLSOS (Figure \ref{FIGQTTY}),
\begin{itemize}
    \item\bm{$13,000\times$}: concentration oscillates along reduction 
    time. This scenario occurs due to the amount of blurred points in 
    photomicrographs corresponding to this magnification (namely 6, 60 
    and 120 minutes); this fact can also be attributed to the small 
    amount of nanoparticles contained in the photomicrograph for the 
    reduction time of 9 minutes. The higher concentration of 
    nanoparticles in 15 minutes is attributed to the evaluation of an 
    area with nanoparticles accumulated on the surface, for better 
    comprehension of size and shape of formed nanoparticles. However, 
    in 30 and 120 minutes it is possible to observe the increase of gold 
    nanoparticles incorporated in the surface of natural rubber, equally 
    presented by UV-Vis spectroscopy results.
    \item\bm{$25,000\times$}: concentration increases logarithmically, 
    tending to stability. This image set presents a coherent behavior 
    when comparing to Figure 8, reaching a plateau at 60 minutes; it is 
    possible to observe in related studies \cite{CABRERA2013B} that 
    nanoparticle growth occurs out of the natural rubber plane from 
    this reduction time, forming aggregates of nanoparticles, 
    difficulting the total nanoparticle evaluation in the natural rubber 
    surface. This difficulty is solved increasing SEM magnification to 
    $200,000\times$. Areas with higher contrast in samples of 9 and 120 
    minutes were labeled with a lower accuracy degree.
    \item\bm{$30,000\times$}: concentration appears to behave as in 
    $25,000\times$, but there is a small oscillation. This contrast 
    difference affects photomicrographs corresponding to 6 and 30 
    minutes. Nanoparticle amount increase after 30 minutes.
    \item\bm{$100,000\times$}: concentration tends to stability after 
    15 minutes. The amount of nanoparticles in 6 minutes appears to be 
    higher than 9 minutes; this situation occurs because surface defects 
    are attributed to nanoparticles through MLSOS. Similarly, blurred 
    areas in 60 and 120 photomicrographs affects the final result.
    \item\bm{$200,000\times$}: concentration behavior resembles that of 
    $100,000\times$. Superficial details were assigned to nanoparticles 
    in photomicrographs of 6 and 9 minutes. Besides, photomicrographs 
    corresponding to 60 and 120 minutes have blurred areas, affecting 
    MLSOS performance.
\end{itemize}

\section{Conclusion}

In this study we consolidate Multi-Level Starlet Segmentation (MLSS) and 
Multi-Level Starlet Optimal Segmentation (MLSOS), techniques presented 
to automatic separate information in photomicrographs. MLSS performs 
separation of areas in several levels, whereas MLSOS chooses the optimal 
segmentation level based on Matthews correlation coefficient (MCC). 
Segmentation reliability is evaluated by precision, recall and accuracy 
measures.

MLSOS is applied to estimate the amount of gold nanoparticles 
incorporated in a natural rubber sample, using five photomicrograph sets 
with different magnifications. The samples were obtained by green 
synthesis of gold nanoparticles using natural rubber membranes, in 
different reduction times. This technique can be used in the 
investigation of chemical kinetics in the synthesis of nanoparticles, 
and also as a complement of UV-Vis spectroscopy, based on image 
datasets.

Segmentation results were compared to UV-Vis spectroscopy, achieving 
satisfactory results mainly in the image set with magnification of 
$25,000\times$. Issues in segmentation are related to surface defects 
being recognized as nanoparticles by MLSOS. However, even in these cases 
MLSOS presents accuracy higher than 88\%.

\section*{Acknowledgments}

The authors would like to acknowledge the Brazilian foundations of 
research assistance CNPq, CAPES and FAPESP. This research is supported 
by FAPESP (Procs 2010/20496-2 and 2011/09438-3).

\end{document}